\documentclass[10pt,twocolumn,letterpaper]{article}

\usepackage{iccv}              % To produce the CAMERA-READY version
\usepackage[accsupp]{axessibility}  % Improves PDF readability for those with disabilities.
\usepackage[T1]{fontenc}
\usepackage{xcolor}
\usepackage{amsmath}
\usepackage{booktabs}
\usepackage{float}
\usepackage{multirow}

\usepackage[most]{tcolorbox}

\usepackage{graphicx}
\usepackage{color}

\usepackage{pifont}  % For uniform tick and cross

\definecolor{iccvblue}{rgb}{0.21,0.49,0.74}
\usepackage[pagebackref,breaklinks,colorlinks,allcolors=iccvblue]{hyperref}

\usepackage{makecell}

\def\paperID{*****} % *** Enter the Paper ID here
\def\confName{ICCV}
\def\confYear{2025}

\title{Towards Reliable and Interpretable Document Question Answering via VLMs}

\author{
Alessio Chen$^*$, Simone Giovannini$^*$, Andrea Gemelli$^+$, Fabio Coppini$^+$, Simone Marinai$^*$\\ \ \\
{\small $(*)$} Università degli Studi di Firenze, Via di Santa Marta 3, Florence, 50139, Italy\\
{\tt\small alessio.chen@edu.unifi.it, simone.giovannini1@unifi.it, simone.marinai@unifi.it}\\
{\small $(+)$} Letxbe.ai, 229 Rue Saint-Honoré, 75001 Paris, France \\
{\tt\small andrea.gemelli@letxbe.ai, fabio.coppini@letxbe.ai}
}

\begin{document}
\maketitle
\begin{abstract}

Vision-Language Models (VLMs) have shown strong capabilities in document understanding, particularly in identifying and extracting textual information from complex documents. Despite this, accurately localizing answers within documents remains a major challenge, limiting both interpretability and real-world applicability. 
To address this, we introduce \textit{DocExplainerV0}, a plug-and-play bounding-box prediction module that decouples answer generation from spatial localization.
This design makes it applicable to existing VLMs, including proprietary systems where fine-tuning is not feasible. Through systematic evaluation, we provide quantitative insights into the gap between textual accuracy and spatial grounding, showing that correct answers often lack reliable localization. Our standardized framework highlights these shortcomings and establishes a benchmark for future research toward more interpretable and robust document information extraction VLMs.

\end{abstract}    
\section{Introduction}
\label{sec:intro}

VLMs are nowadays recognized as the SOTA models for most of the multi-modal understanding tasks \cite{vision-language-survey-2024, Docfomerv2-2023, molmo-2024} and notably in document (visual) question answering (Document VQA) \cite{DOCVQA-2021, DUDE-2023, DOCMATRIX-2024, BigDocs-2025}, where the model is prompt with the image of a document and a question about it. Recent generative models often provide correct answers, yet they rarely correctly indicate where those answers are located in the document \cite{WHERE-IS-COME-FROM-2025, nourbakhsh-etal-2024-towards, LEARNINGTOGROUND-2024}. In other words, while textual accuracy is high, spatial grounding remains unreliable.
This limitation arises from a fundamental mismatch in training objectives: generative VLMs are optimized for autoregressive next-token prediction, whereas classical vision models rely on explicit spatial supervision, such as bounding-box regression \cite{jiao2025from, LayTextLLM-2024, wang-etal-2024-docllm, FoundationalVLMsLimitations-2025}.
To the authors' knowledge, the only prior attempt to analyze spatial grounding in document VQA is \cite{DocVXQA-2025}. Beyond that, there is no systematic, quantitative evaluation. In this work, we address this gap by using the BoundingDocs dataset\cite{BoundingDocs-2025} to provide the first comparable and quantitative analysis of the ability of VLMs to localize answers within documents.

% As we are not interested in training a model to extract both the token and its position (this has already been done in the literature \cite{LayoutLMv3-2022, SMOLDOCLING-2025, LMDX-2024, PIN-2024,BoundingDocs-2025}), our attention will firstly be focused on benchmarking existing generative models, without any pre-training or fine-tuning, yet using different prompts. By comparing different prompting techniques, we investigate how current VLMs "reason" and perform with regards to the spatial position of the answer; furthermore, we propose a simple architecture (DocExplainerV0) as a starting point to ground VLMs answers with spatial truth. Finally, we compare all these approaches with a naive algorithm build on top of an existing OCR (Textract \cite{aws_textract}), as well as a classical commercial solution (Claude Sonnet 4 \cite{anthropic_claude_sonnet4_system_card}), to benchmark very large models and provide a comprehensive analysis. Maybe surprisingly, the naive algorithm is the one getting the best results, showing that the current models, independently of their size, are not yet suited for precise information extraction.

Instead of training models to jointly predict answers and their positions (as explored in prior works \cite{LayoutLMv3-2022, SMOLDOCLING-2025, LMDX-2024, PIN-2024,BoundingDocs-2025}), we focus on evaluating existing generative VLMs in scenarios where training is impractical or impossible, such as with proprietary models or very large pre-trained systems. Our main contributions are: 

\begin{itemize}
    \item \textbf{DocExplainerV0} \footnote{\url{https://huggingface.co/letxbe/DocExplainer}}: A plug-and-play modular architecture to spatially ground VLM answers without any retraining, enabling existing models to provide interpretable outputs.
    \item \textbf{Benchmarking and Quantitative Analysis \footnote{\url{https://github.com/letxbe/research/tree/main/doc-explainer}}}: Using the BoundingDocs dataset \cite{BoundingDocs-2025}, we provide a systematic and comparable evaluation of how well VLMs localize answers within documents.
    \item \textbf{Research Questions}: Our work addresses key questions, including:
    \begin{enumerate}
        \item How well do current generative VLMs localize answers relative to their textual accuracy?
        \item How can modular or hybrid approaches (like decoupling answer generation and localization) address the spatial grounding gap?
        \item How can VLM interpretability be quantified for document VQA beyond textual accuracy?
    \end{enumerate}
\end{itemize}

In Section \ref{sec:related_works}, we review related work on document understanding, spatial grounding, and interpretability in VLMs. Section \ref{sec:methods} describes our methodology, including model selection, prompting strategies, and the DocExplainerV0 architecture. Section \ref{sec:experiments} presents the experimental setup, dataset details, evaluation metrics and results. Finally, Section \ref{sec:conclusions} concludes the paper and outlines directions for future research.

\section{Related Works}
\label{sec:related_works}

% In this section we review the state of the art, starting from existing solutions on Document Understanding and then [...continue]
VLMs have become the dominant approach for document understanding, often combining OCR-extracted text with visual features to answer questions \cite{molmo-2024, wang-etal-2024-docllm, Docfomerv2-2023, DOCVLM-2024}. These OCR-based models leverage rich linguistic context, which supports strong reasoning abilities, but their reliance on text extraction introduces vulnerabilities: misread characters, missing text, or layout errors can propagate through the system \cite{OCRImpact-2025}. OCR-free alternatives process document images directly with vision encoders specialized pretraining objectives \cite{Donut-2021, Pix2Struc-2023, DLAVA-2024}. While promising, such models typically require task-specific fine-tuning and still struggle to achieve strong zero-shot performance. \cite{TEXTVQA-2019, CHARTQA-2022, DCOVQA-2021, DOCMATRIX-2024, BigDocs-2025}.

Interpretability and spatial grounding have attracted growing interest, with methods aiming to highlight relevant regions, generate attribution maps, or produce visual explanations \cite{DocXplain-2024, DLAVA-2024, PixelSHAP-2025}. In document VQA specifically, DocVXQA \cite{DocVXQA-2025} generates heatmaps to indicate ares associated with answers, but these explanations remain qualitative, and no widely accepted metric exists for quantitative evaluation. 
Spatial localization techniques fall into two broad categories: prompt-based strategies that encode positional information without retraining \cite{LLMWrapper-2025, LMDX-2024, DLAVA-2024}, and architecture-level methods that integrate layout-aware components for more precise grounding \cite{PIN-2024, LayTextLLM-2024}. Prompt-based approaches are lightweight but inconsistent, while architectural solutions require retraining and higher model complexity.
Despite these efforts, the document understanding literature still lacks a standardized, quantitative benchmark for evaluating how well VLMs localize answers.

\section{Methods}
\label{sec:methods}

\begin{figure*}[t]
\centering
\includegraphics[width=1\linewidth]{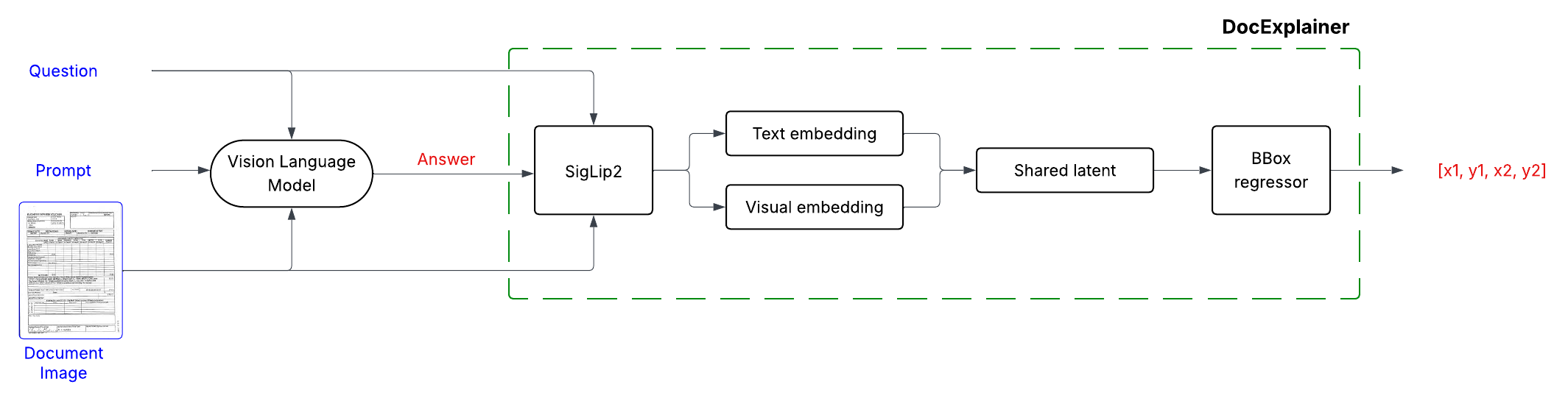}
\caption{Architecture of the learned bounding box predictor. Visual and textual embeddings are fused in a shared latent space and passed through fully connected layers to produce bounding box coordinates.}
\label{fig:bbox_predictor}
\end{figure*}

\subsection{Model choice and Prompting}

We focus our evaluation on three representative families of recent VLMs, chosen to cover different scales and design philosophies.

\begin{itemize}
    \item \textbf{SmolVLM-2.2B} \cite{SMOLVLM-2025}: a lightweight open-source VLM (2.2B parameters), optimized for efficiency and small-scale deployment. This serves as a lower-bound reference for compact models on document VQA.
    \item \textbf{QwenVL2.5-7B} \cite{QWEN2.5VL-2025}: a mid-sized open-source model that demonstrates strong performance across multimodal benchmarks, representing a balance between scalability and accessibility for research.
    \item \textbf{Claude Sonnet 4} \cite{anthropic_claude_sonnet4_system_card}: a proprietary large-scale model with competitive multimodal capabilities.
    
\end{itemize}
Three prompting strategies were tested for all open-source VLMs:
\begin{itemize}
    \item \textbf{Zero-shot}: The model receives only the question and the document image without additional context, as shown in the example in Figure \ref{fig:prompt_template}.
    
    \item \textbf{Chain-of-Thought (CoT)}: The zero-shot prompt is additionally enriched with QA pairs, encouraging step-by-step reasoning. 
    
    \textit{Example:} \texttt{Q: "What is the invoice date?" A: \{"value":"2025-08-19", "position": [100, 50, 200, 30]\}}
    
    \item \textbf{Anchor-based prompting}: The OCR-extracted words provided by BoundingDocs are used as anchors to enhance the zero-shot prompt by providing the location of words on the page. 
    
    \textit{Example:} \texttt{The word "Invoice" is at [50, 20, 80, 20]}
\end{itemize}
The commercial VLM was evaluated in the zero-shot setting using the best prompting strategy resulting from the previous tests.

In every case, the VLM is instructed to return prediction in a JSON format containing both the answer and its bounding box, as shown in Figure \ref{fig:prompt_template}. This enables standardized computation for both textual and spatial metrics.

\iffalse
\begin{figure*}[h!]
\centering
\begin{tabular}{ccc}
    \includegraphics[width=0.32\textwidth]{images/cot.png} &
    \includegraphics[width=0.32\textwidth]{images/anchors.png} &
    \includegraphics[width=0.32\textwidth]{images/zero_shot.png} 
\end{tabular}
\caption{Example of the three prompting strategies used.The predicted bounding boxes are requested in the [0,1000] range to match the BoundingDocs annotations.}
\label{fig:prompting_strategies}
\end{figure*}
\fi

\begin{figure}[htbp]
\centering
\begin{tcolorbox}[
    colback=blue!5,
    colframe=blue!40!black,
    boxrule=0.8pt,
    arc=2pt,
    left=6pt,
    right=6pt,
    top=6pt,
    bottom=6pt,
    width=\columnwidth
]
{\footnotesize
Based only on the document image, answer the following question:\\[0.2em]
\textbf{Question:} What is the total amount due?\\[0.3em]
Provide \textbf{ONLY} a JSON response in the following format:
\begin{verbatim}
{
  "content": "answer",
  "position": [x, y, w, h]
}
\end{verbatim}
Each position value \textbf{MUST} be in the range [0, 1000].
}
\end{tcolorbox}
\caption{Zero-shot prompt example used to query the VLMs.}
\label{fig:prompt_template}
\end{figure}

\subsection{DocExplainerV0}
In addition to the prompting strategies, we evaluate a model architecture that decouples answer prediction from spatial localization. In this setup, a VLM generates the answer text, while a dedicated bounding box predictor, built on top of the SigLiP2 \cite{SigLip2-2025} vision and text backbone, estimates the corresponding position within the document. 

The bounding box regressor adopts a dual-branch design: visual and textual embeddings from SigLiP2 are first projected into a shared latent space and then fused through fully connected layers, with a regression head producing the final bounding box coordinates: $[x_1,y_1,x_2,y_2]$. During training, the SigLiP2 backbone remains frozen, and only the bounding box regressor is updated. The model is optimized using the Smooth L1 loss (Huber loss) on normalized bounding box coordinates in [0,1], defined as:
\[
L(x,y) =
\begin{cases} 
0.5 (x-y)^2 & \text{if } |x-y| < 1, \\
|x-y| - 0.5 & \text{otherwise}.
\end{cases}
\]

Training is conducted for 20 epochs on the training split of BoundingDocs v2.0, using a single NVIDIA L40S-1-48G GPU. The final checkpoint is selected based on the highest mean IoU score achieved on the validation split. An overview of the bounding box predictor architecture is shown in Figure \ref{fig:bbox_predictor}.

\iffalse
\begin{figure}[h!]
\centering
\includegraphics[width=1\linewidth]{images/bbox_predictor.png}
\caption{Architecture of the BoundingDocsv0. Visual and textual embeddings are fused in a shared latent space and passed through fully connected layers to produce bounding box coordinates.}
\label{fig:bbox_predictor}
\end{figure}
\fi

\section{Experiments}
\label{sec:experiments}

\begin{table}[ht!]
\centering
\begin{tabular}{l c c c}
\hline
\textbf{Architecture} & \textbf{Prompting} & \textbf{ANLS} & \textbf{MeanIoU} \\
\hline
\multirow{3}{*}{Smol} 
  & Zero-shot   & .527 & .011 \\
  & Anchors     & .543 & .026 \\
  & CoT         & .561 & .011 \\
\hline
\multirow{3}{*}{Qwen} 
  & Zero-shot   & .691 & .048 \\
  & Anchors     & .694 & .051 \\
  & CoT         & \underline{.720} & .038 \\
\hline
Claude Sonnet 4 & Zero-shot & \textbf{.737} & .031 \\
\hline
\multirow{2}{*}{\begin{tabular}[c]{@{}l@{}}Smol + D.E.\\ Qwen + D.E.\end{tabular}} 
  & \multirow{2}{*}{Zero-shot} & .572 & .175 \\
  &                            & .689 & .188 \\
\hline
\multirow{2}{*}{\begin{tabular}[c]{@{}l@{}}Smol + Naive OCR\\ Qwen + Naive OCR\end{tabular}} 
  & \multirow{2}{*}{Zero-shot} & .556 & \underline{.405} \\
  &                            & .690 & \textbf{.494} \\

\hline
\end{tabular}\caption{Document VQA performance of different models and prompting strategies on the BoundingDocs v2.0 dataset. Smol stands for SmolVLM-2.2B, Qwen stands for Qwen2-VL-7B, and D.E. stands for DocExplainerV0. The best value is shown in \textbf{bold}, the second-best value is \underline{underlined}.}
\label{tab:results}
\end{table}

\begin{figure*}[ht!]
\centering
\begin{tabular}{ccc}
    \includegraphics[width=0.22\textwidth]{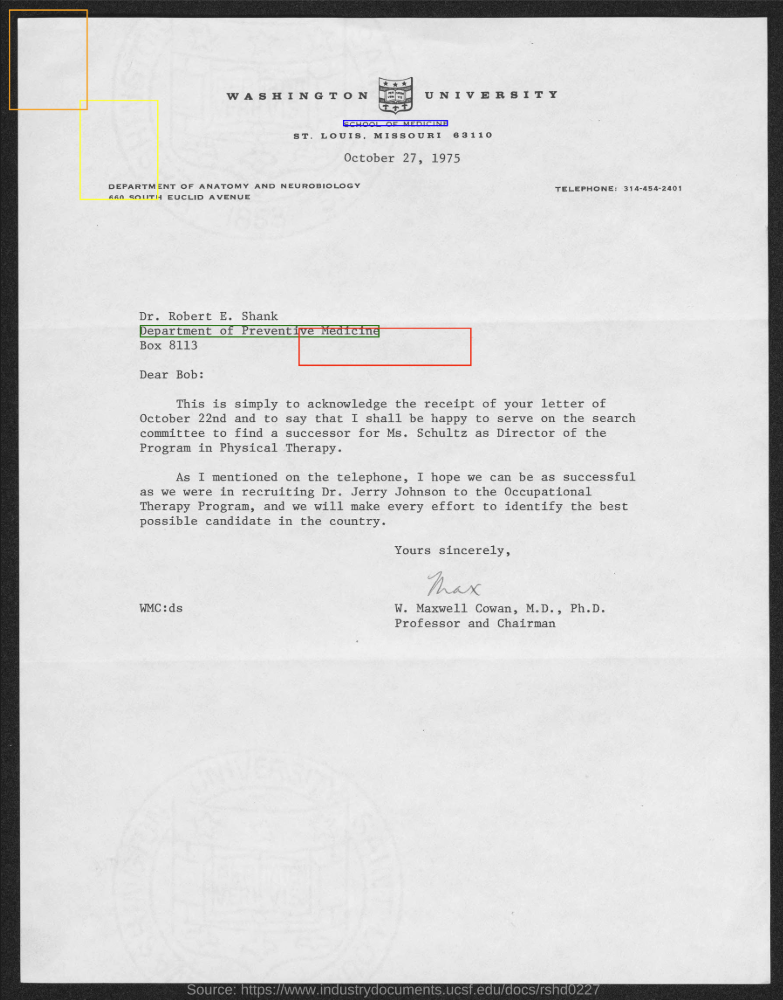} &
    \fbox{\includegraphics[width=0.22\textwidth]{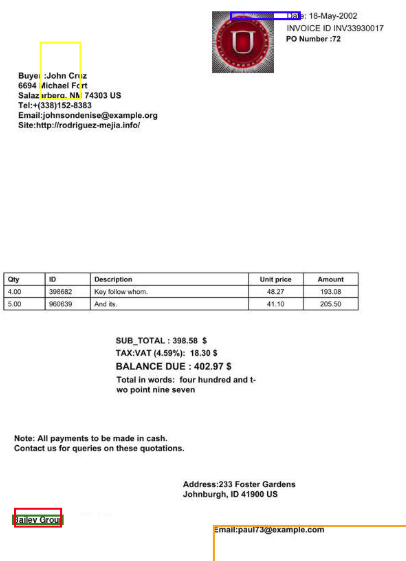}} &
    \includegraphics[width=0.22\linewidth]{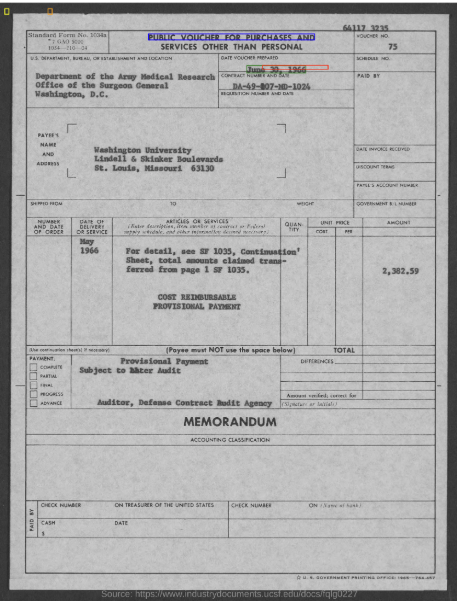} \\

    \makecell{Q: "What is the department \\ of Dr. Robert?" \\ 
              A: "Department of Preventive Medicine"} &
    \makecell{Q: "What is the name \\ of the seller?" \\ 
              A: "Bailey Group"} &
    \makecell{Q: "What is the date \\ voucher prepared?" \\ 
              A: "June 30, 1966"} \\
\end{tabular}
\caption{Example document images from BoundingDocs with predicted bounding boxes overlaid. 
The bounding box colors indicate: 
\textcolor{green}{\rule{0.2cm}{0.2cm}} Ground truth, 
\textcolor{red}{\rule{0.2cm}{0.2cm}} DocExplainer, 
\textcolor{orange}{\rule{0.2cm}{0.2cm}} CoT prompting, 
\textcolor{yellow}{\rule{0.2cm}{0.2cm}} Zero-shot prompting, 
\textcolor{blue}{\rule{0.2cm}{0.2cm}} Anchor-based prompting. 
The bounding boxes from prompting-based methods are all far from the ground truth, while the only one close to it is DocExplainer.}
\label{fig:bbox_examples}
\end{figure*}

\subsection{Dataset}

We evaluate on BoundingDocs v2.0\footnote{\url{https://huggingface.co/datasets/letxbe/BoundingDocs}}, which extends BoundingDocs \cite{BoundingDocs-2025}. The dataset consolidates 11 public sources, including invoices, contracts, forms, receipts, and multilingual corpora, into a unified QA format with normalized bounding boxes for precise answer localization. It comprises 48{,}151 documents (237{,}437 pages) and 249{,}016 question–answer pairs across eight languages, enabling evaluation of both answer correctness and spatial grounding.

Relative to the initial release \cite{BoundingDocs-2025}, v2.0 delivers targeted improvements in consistency, linguistic coverage, and data quality:
\begin{itemize}
\item \textbf{Universal Question Rephrasing:} Every entry now includes a \texttt{rephrased\_question} field; for human-written questions this field mirrors \texttt{question} to standardize the schema.
\item \textbf{Expanded Multilingual Coverage:} XFUND questions are rephrased using Claude 3.7 Haiku, improving cross-language consistency.
\item \textbf{Data Alignment Correction:} A mismatch between \texttt{doc\_images} and \texttt{doc\_ocr} in some MP-DocVQA entries has been fixed, ensuring image–OCR alignment.
\end{itemize}

\subsection{Metrics and Evaluation}

We evaluate model performance with two metrics: textual accuracy and spatial localization.  

\paragraph{Textual Accuracy.}  
We use the Average Normalized Levenshtein Similarity (ANLS) \cite{ANLS*-2025}, which measures the normalized Levenshtein distance between predicted and ground-truth answers. ANLS is computed for each QA pair and then averaged over all samples in the test set. It ranges from 0 (no overlap) to 1 (exact match), and is tolerant to minor spelling or formatting variations.

\paragraph{Spatial Localization.}  

To evaluate localization, we compute the standard Intersection over Union (IoU) value between predicted bounding boxes $B_\text{pred}$ and reference boxes $B_\text{gt}$ areas:
\[
\text{IoU} = \frac{|B_\text{pred} \cap B_\text{gt}|}{|B_\text{pred} \cup B_\text{gt}|}.
\]
The MeanIoU is obtained by averaging the IoU scores across all QA pairs in the test set, providing a single measure of spatial accuracy.

\subsection{Naive OCR-based Baseline}
As a baseline for answer localization, we implement a simple OCR-based method. Specifically, we query the VLM for an answer and then search for this predicted value within the OCR text extracted from the document (provided by the BoundingDocs dataset). When a match is found, the corresponding bounding box in the page is returned as the predicted location. If the complete answer cannot be matched in the OCR text, we instead use the first word of the VLM’s answer to perform the same procedure.

\subsection{Results}

The results of our experiments are presented quantitatively in Table \ref{tab:results} and qualitatively in Figure \ref{fig:bbox_examples}.  
We can notice that VLMs are unable to provide the position of the value extracted from the document, and return completely incorrect bounding boxes, as demonstrated by the IoU values. 
Among the various prompting settings, "Anchors" is the one that most helps VLMs on this task, but without bringing significant advantages, while still compromising ANLS values, whose highest scores are achieved in the "CoT" setting.

The use of DocExplainer instead significantly improves the MeanIoU values of both models, demonstrating the effectiveness of the proposed architecture. It is nonetheless fundamental to emphasize that DocExplainer's results are still very far from the OCR-based baseline (approximately three times less effective). This is also due to the type of data used, since in BoundingDocs the answers always have an exact match in the document and therefore OCR search performs well. 
However, our objective is to propose an architecture that finds the relevant area in the document even for questions whose answer is not  explicitly in the text, a case in which the OCR baseline would obviously fail.

\section{Conclusion}
\label{sec:conclusions}
In this work we investigate the spatial grounding capabilities of VLMs (both open and closed weights) for document understanding and compare it with an OCR-based method.
Rather than providing a complete solution, our goal was to establish a benchmark for future research and highlight the challenges inherent in document-level spatial reasoning. Importantly, this study represents one of the first attempts to provide quantitative interpretability in document VQA by evaluating not only the correctness of answers but also their spatial grounding.
Using the BoundingDocs dataset, we restricted our analysis to questions where the answer can be clearly localized within a single bounding box, excluding cases that require reasoning over multiple elements or regions. While naive OCR-based methods provide a useful upper bound for spatial grounding, they also highlight a limitation of the dataset: many answers are not fully captured by OCR, and multiple regions may contribute to understanding a given answer, which cannot be properly evaluated under current single-box annotations.

% Notably, simple OCR-based baselines often achieve comparable results, emphasizing the need for careful and fair comparisons when evaluating spatial grounding methods.

% Prompting strategies are included primarily as an experimental setting to maximize the performance of existing VLMs without retraining and to highlight their limitations quantitatively. Similarly, comparisons with a naive OCR pipeline (Textract) and a commercial VLM solution (Claude Sonnet 4 \cite{anthropic_claude_sonnet4_system_card}) are intended to provide context, not as a proposed solution. While naive OCR sometimes achieves higher spatial scores on the dataset, this outcome is expected and reflects a bias: some questions in BoundingDocs do not have ground-truth annotations in the OCR layer. In broader real-world scenarios, such naive methods would fail to provide reliable grounding. This limitation and its implications for generalization are discussed further in the limitations and conclusion sections.

{
\small
 \section*{Acknowledgments}
 We thank the CAI4DSA actions (Collaborative Explainable neuro-symbolic AI for Decision Support Assistant) of the FAIR national project on artificial intelligence, PE1 PNRR (https://fondazione-fair.it/).

\bibliographystyle{ieeenat_fullname}
\bibliography{sample}
}

\end{document}